\documentclass[11pt]{article}
\usepackage{acl2014}
\usepackage{times}
\usepackage{url}
\usepackage{latexsym}
\usepackage{microtype}		
\usepackage{graphicx}
\usepackage{multirow}
\usepackage{cleveref}
\usepackage{amsmath}
\usepackage{amssymb}
\usepackage{floatrow}
\usepackage{caption}
\usepackage{tikz-dependency}
\usepackage{tikz}

\title{Targetable Named Entity Recognition in Social Media}

\author{Sandeep Ashwini \\
  Amazon Fulfillment Technologies \\
  Amazon.com, LLC \\
  Seattle, WA 98109 \\
  {\tt asandee@amazon.com} \\\And
  Jinho D.\ Choi \\
  Department of Computer Science\\
  Emory University \\
  Atlanta, GA 30322 \\
  {\tt choi@mathcs.emory.edu} \\}

\date{}

\begin{document}
\maketitle
\begin{abstract}
We present a novel approach for recognizing what we call targetable named entities; that is, named entities in a targeted set (e.g, movies, books, \textsc{tv} shows).
Unlike many other NER systems that need to retrain their statistical models as new entities arrive, our approach does not require such retraining, which makes it more adaptable for types of entities  that are frequently updated.
For this preliminary study, we focus on one entity type, movie title, using data collected from Twitter.
Our system is tested on two evaluation sets, one including only entities corresponding to movies in our training set, and the other excluding any of those entities.
Our final model shows F1-scores of 76.19\% and 78.70\% on these evaluation sets, which gives strong evidence that our approach is completely unbiased to any particular set of entities found during training.
\end{abstract}


\section{Introduction}

Current state-of-the-art systems perform very well on recognizing named entities such as persons, locations, or organizations~\cite{lin:09a,ratinov:09a,turian:10a}; however, there is relatively little work on recognizing other types of entities such as movies, books, songs, or \textsc{tv} shows~\cite{ritter:11a,guo:13a}.
As more people share information about various topics, named entity recognition for these non-traditional entities becomes more important; especially using data collected from social media, where people's interests are directly expressed.

Even without the difficulties of processing data from social media~\cite{foster:11a,dent:11a}, NER for non-traditional entities can be more challenging than traditional ones.
First, it is hard to build a large dataset containing enough instances of these entities, which is required by most state-of-the-art systems using statistical machine learning.
Previous studies have shown that these types of entities comprise only a small portion of the entire data consisting of random streams~\cite{ritter:11a,gattani:13a}, which is not sufficient to build a good statistical model.
Thus, a more focused dataset needs to be constructed to build a good statistical model for these entities.

Second, non-traditional named entities tend to be updated more frequently than traditional ones (e.g., new movies are released every week), which makes it difficult to build a statistical model that performs well for a long period of time without being retrained.
It is possible to keep annotating more data for newly introduced entities~\cite{finin:10a} and retraining new models accordingly; however, the choice of which data to annotate may not be always clear and retraining a new model is not necessarily easy, especially for people bound to use off-the-shelf systems.
Thus, it is desirable to build a system that can be dynamically adjusted for new entities without having to be retrained.

Third, traditional named entities mostly occur in the form of noun phrases, whereas non-traditional ones do not necessarily follow the same trend (e.g., movie titles such as \textit{Frozen}, \textit{Ride Along}), which makes them more difficult to recognize.
One bright side is that there exists a closed set containing some of these entity types such that the entire space is already known for certain types although it may expand quickly.
Thus, even though non-traditional named entities can come in various forms, a system can anticipate the general forms they take given the closed set.
Throughout this paper, we refer to these types of entities as ``Targetable Named Entity''; that is, named entities in a targeted set that is closed but can be updated (e.g., a set of movie titles).
A general NER system may still perform well for targetable named entities; however, a more focused system can be designed to give similar performance when the entity set is assumed.

\begin{table*}[htbp!]
\captionsetup{width=0.85\textwidth}
\centering
\begin{tabular}{c|c||c|c||c|c}
\textbf{Movie} & \textbf{Count} &\textbf{Movie} & \textbf{Count} &\textbf{Movie} & \textbf{Count}\\ 
\hline\hline
Hobbit             & 223 & Non Stop      & 29 & Son of God        & 15 \\
Frozen             & 131 & Lego Movie    & 22 & American Hustle   & 14 \\
Gravity            & 126 & Lone Survivor & 20 & Ride Along        & 14 \\
12 Years a Slave   & 115 & Nebraska	     & 16 & Lord of the Rings & 13 \\
Dallas Buyers Club &  41 & Man of Steel  & 15 & Titanic           &  7 \\
\end{tabular} 
\caption{\protect Distribution of the top 15 most frequent movies in our data.  The count columns show the number of entities annotated for each movie.}
\label{tbl:movie-distribution}
\end{table*}

\noindent In this paper, we give a careful analysis of NER for one targetable named entity type, movie title, using data collected from Twitter.
We begin by describing our newly introduced corpus used for developing and evaluating our approach (\Cref{sec:corpus}).
We then present a three-stage NER system, where each stage performs normalization, candidate identification, and entity classification, respectively (\Cref{sec:approach}).
Finally, we evaluate our approach on two datasets, one including only entities corresponding to movies in our training set, and the other excluding any of those entities (\Cref{sec:experiments}).

To the best of our knowledge, this is the first time that a targetable named entity type is thoroughly examined.
Our approach is general enough to be applied to other targetable named entity types such as \textsc{tv} show, song title, product name, etc.


\section{Corpus}
\label{sec:corpus}

\subsection{Data collection}

We collected data from Twitter because it was one of the fastest growing social platforms in terms of both users and documents~\cite{smith:13a}. Instead of crawling random tweets, we collected tweets filtered by specific keywords using the Tweepy API.\footnote{\url{http://www.tweepy.org}} These keywords correspond to movies that were popular during the collection time period, between Dec.\ 2013 and Feb.\ 2014 (e.g., \textit{Hobbit}, \textit{Frozen}).
It was difficult to find tweets including older movies; however, we managed to collect a few tweets corresponding to movies prior to that time period (e.g., \textit{Lord of the Rings}, \textit{Titanic}).
\Cref{tbl:movie-distribution} shows the distribution of the most frequent movies in our data.

We targeted tweets including specific keywords instead of general tweets for two reasons. 
One is that targeted streams are more commonly used by commercial organizations for monitoring response to their products~\cite{li:12a}.
We expect that interested parties would find this type of data collection more useful than a larger body of random tweets.
The other is that the proportion of tweets including movie entities among random tweets is usually too small to build a good statistical model and give a meaningful evaluation.
For example, \newcite{ritter:11a} annotated 2,400 random tweets with 10 entity types; however, only 34 movie entities were found in that data, which made it difficult for them to show how well their system would perform for this entity type.
By restricting the search to a single entity type, we can collect more focused data, which leads to better modeling and evaluation for that entity type.

All redundant tweets and tweets not written in English are removed from our corpus.
As a result, a total of 1,096 tweets were collected, of which, 71.9\% include at least one movie entity.

\subsection{Annotation}
\label{ssec:annotation}

\begin{table*}[htbp!]
\captionsetup{width=1.0\textwidth}
\centering
\begin{tabular}{c|l|c}
\# & \multicolumn{1}{|c|}{\textbf{Example}} & \textbf{Note}\\
\hline\hline
\multirow{3}{*}{1} & Can't wait to see \textbf{the \#Hobbit} tomorrow!                  & Main Title\\
                   & You better believe "\textbf{the Desolation of Smaug}," was badass! & Sub Title \\
                   & I really like \textbf{the Hobbit 2: The Desolation of Smaug}.      & Full Title\\
\hline
\multirow{2}{*}{2} & How many times can you watch \textbf{Despicable me 2}? & Sequel \#: 2\\
                   & Going to see \textbf{\#Hobbit 2} this afternoon.       & Sequel \#: 2\\
\hline 
\multirow{2}{*}{3} & Dedicated 3 hours of my life to \textbf{the wolf on wall street} last night... & on $\rightarrow$ of \\
                   & Finally got \textbf{12 Years as a Slave} on Blu-Ray!                           & as $\rightarrow \emptyset$ \\
\hline
\multirow{2}{*}{4} & I literally just watched her in \textbf{12Yrs} on Saturday.  & 12Yrs $\rightarrow$ 12 Years a Slave \\
                   & I'm gonna have a \textbf{Hobbit} \& \textbf{LoT} marathon... & LoT   $\rightarrow$ Lord of the Rings \\
\hline
\multirow{2}{*}{5} & \multicolumn{2}{|l}{Not sure between \textbf{\#hobbit}, \textbf{\#anchorman2}, or \textbf{\#americanhustle} this weekend?}\\ 
                   & \multicolumn{2}{|l}{Gonna see \textbf{\#DesolationOfSmaug} finally!!! In 3D! OMG, \#uncontrollablyexcited!!}\\ 
\end{tabular} 
\caption{\protect Examples for each guideline in \Cref{ssec:annotation}, where annotation is shown in bold.  When a hashtag is used to indicate a movie title, it is considered a part of the entity (e.g., the \#Hobbit).}
\label{tbl:annotation-example}
\end{table*}

The following are the annotation guidelines used to label movie entities in our corpus.
\Cref{tbl:annotation-example} shows annotation examples for each guideline.

\begin{enumerate}
\item \textbf{Title match}: Each sequence of tokens matching a main title (e.g., \textit{the \#Hobbit}), a sub title (e.g., \textit{the Desolation of Smaug}), or a full title (e.g., \textit{the Hobbit 2: The Desolation of Smaug}) is annotated as a single entity.
\item \textbf{Sequel match}: Sequel numbers are considered part of entities (e.g., \textit{Despicable me 2}).
\item \textbf{Typo match}: Entities including typos are still annotated (e.g., \textit{the wolf on wall street}, which should be \textit{the wolf of wall street}).
\item \textbf{Abbreviation match}: Abbreviations indicating movie titles are annotated (e.g., \textit{12Yrs}, which should be \textit{12 Years a Slave}).
\item \textbf{Hashtag match}: Individual hashtags used to indicate movie titles are annotated (e.g., \textit{\#DesolationOfSmaug} is annotated, whereas \textit{\#uncontrollablyexcited} is not).
\end{enumerate}

\noindent All movie entities were manually annotated by a graduate student.
A total of 911 movie entities were annotated, corresponding to 53 different movies.
Note that a single tweet can include zero to many movie entities.
Although our corpus is single annotated, it is done by the author of the guidelines such that the quality of our corpus is most likely higher than ones achieved through crowdsourcing.
Once we show the feasibility of our approach, we plan to increase the volume of our corpus through crowdsourcing~\cite{finin:10a} and also include more entity types.
Our corpus is publicly available: {\footnotesize https://github.com/sandeepAshwini/TwitterMovieData}.


\section{Approach}
\label{sec:approach}

\begin{figure}[htbp!]
\centering
\includegraphics[scale=0.6]{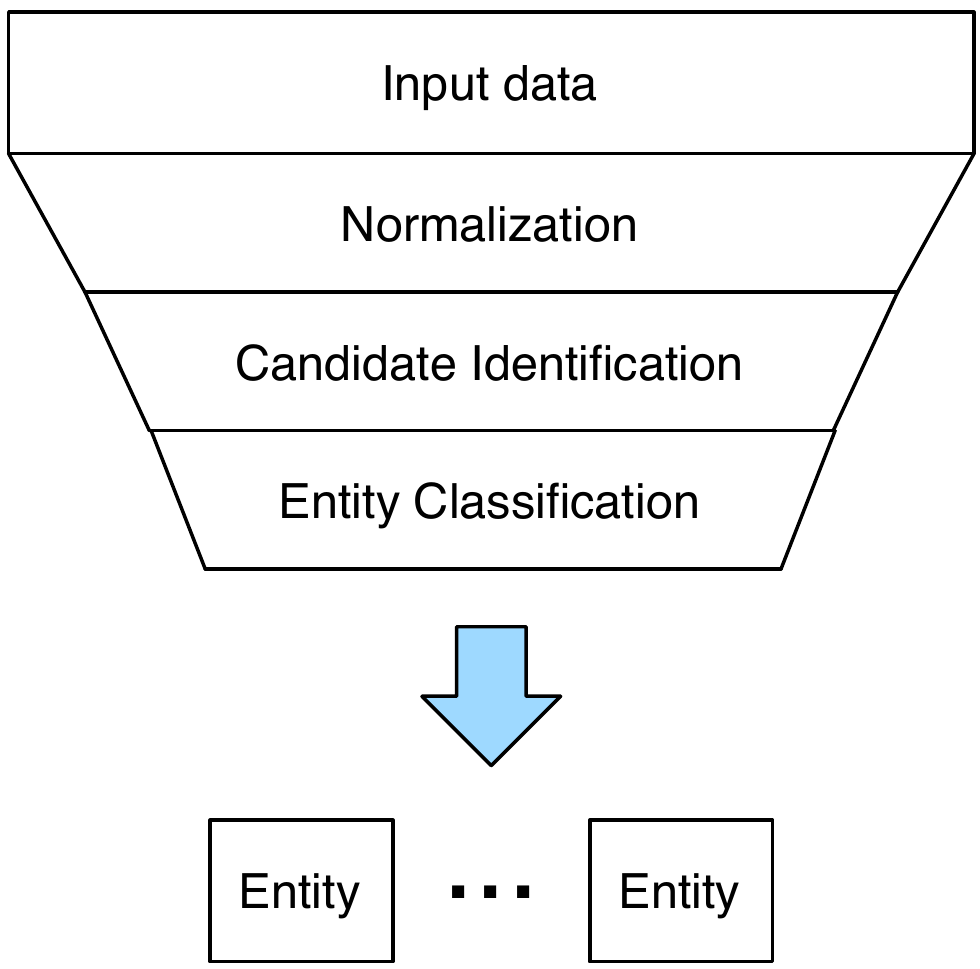}
\caption{The architecture of our NER system.}
\label{fig:overview}
\end{figure}

\Cref{fig:overview} shows the architecture of our three-stage named entity recognition system, which should be general enough to handle most targetable named entity types.
Given input data, a single tweet in our case, tokens are first normalized (either removed or substituted with generic tokens; \Cref{ssec:normalization}).
From the normalized tweet, candidate entities are identified by using a pre-defined gazetteer and heuristics (\Cref{ssec:candidate-identification}).
Finally, each candidate is classified as either an entity or not using a statistical model trained on various features (\Cref{ssec:entity-classification}).


\subsection{Normalization}
\label{ssec:normalization}

Some tokens in tweets act rather like noise.
For instance, hyperlinks, appearing in almost every tweet, usually introduce noise to the NER process and it is hence better to
discard them.
On the other hand, it is better to segment hashtags consisting of multiple words when they are syntactically relevant to the context in the tweet.
Thus, it is preferable to normalize tweets before attempting candidate identification and entity classification.
The followings show our normalization rules.
The first rule is general to any social media data whereas the other rules are specific to tweets.

\begin{enumerate}
\item Hyperlinks, non UTF-8 characters, punctuation marks, and English articles are removed.
\item Retweet markers and the following user IDs are removed (e.g, ``RT @eurweb''). 
      Moreover, User IDs are substituted with a generic token (e.g, @JMussehl $\rightarrow$ \textsc{\#user\#}).
\item Hashtags that are not syntactically relevant to the context are removed.  The relevance is determined by \newcite{kaufmann:10a}'s heuristic test.
      Hashtags that are syntactically relevant to the context are segmented by matching them against tokens in our gazetteer (\Cref{sssec:gazetteer}) using dynamic programming.
\end{enumerate}

\begin{table*}[htbp!]
\centering
\begin{tabular}{c|l}
\# & \multicolumn{1}{|c}{\textbf{Example}}\\
\hline\hline
\multirow{2}{*}{0} & RT @eurweb The I bus i was on, was on fire and and I saw The Hobbit: 2 with @JMussehl!\\
                   & and it was so \#damngood! \#Hobbit \#busfire http://t.co/M2nYROTpdS \\ 
\hline 
\multirow{2}{*}{1} & RT @eurweb I bus i was on was on fire and and I saw Hobbit 2 with @JMussehl\\
                   & and it was so \#damngood \#Hobbit \#busfire \\ 
\hline 
\multirow{2}{*}{2} & I bus i was on was on fire and and I saw Hobbit 2 with \textsc{\#user\#}\\
                   & and it was so \#damngood \#Hobbit \#busfire \\ 
\hline 
\multirow{2}{*}{3} & I bus i was on was on fire and and I saw Hobbit 2 with \textsc{\#user\#}\\
                   & and it was so \#damn \#good\\ 
\end{tabular} 
\caption{An example of our normalization process by each rule in \Cref{ssec:normalization}.}
\label{tbl:normalization-example}
\end{table*}

\noindent After running through these rules, tweets are in their normalized forms, where only relevant information is preserved.
Normalization helps speed up the candidate identification process by reducing the total number of tokens in the tweet; in our data, it gives a 22.87\% reduction in the number of tokens.
Moreover, normalization helps to generate more relevant features for the entity classification stage, as much of the noise in the tweet is effectively removed and some sparsity issues are handled.
It is possible to perform further normalization~\cite{li:12a}, which we will explore in the future.


\subsection{Candidate identification}
\label{ssec:candidate-identification}

From a normalized tweet, candidate entities are identified by matching token sequences in the tweet to entries in a gazetteer.
This stage is designed to give high recall while keeping reasonable precision for NER.
We separate this stage from entity classification (\Cref{ssec:entity-classification}) because we do not want the statistical model used for entity classification to be retrained every time the gazetteer is updated.
This way, candidate identification finds candidates that are likely to be entities using the most up-to-date gazetteer while entity classification ascertains whether or not those candidates are valid without being affected by the gazetteer.
We find this approach effective for targetable named entities.

\subsubsection{Gazetteer}
\label{sssec:gazetteer}

We downloaded the entire list of film entities from Freebase~\cite{bollacker:08a} using the Freebase API.\footnote{https://developers.google.com/freebase/}
This gazetteer can be dynamically updated to maintain coverage for latest movies.
Additionally, we collected the release years of all movies in the gazetteer, which are used as features for entity classification (\Cref{sssec:orthographic-features}).

\subsubsection{Sequence matching}
\label{sssec:sequence-matching}

For each token in the tweet, we check if there exists any sequence of tokens, beginning from that token, that matches an entry in our gazetteer.
If a match is found, this token sequence is added to the list of possible candidates for that tweet.
If more than one matches are found, only the longest sequence is added to the list.
Some previous studies restricted this matching to a $k$-number of tokens to reduce complexity~\cite{guo:13a}.
Instead of limiting the sequence matching to a fixed size, we adapted an affix tree~\cite{weiner:73a}, which tremendously reduced the matching complexity.\footnote{Traversing through the entire gazetteer takes $O(\log |W|)$, where $W$ is a set of all token types in the gazetteer.}
Thus, such restriction was no longer necessary in our system.

Four types of sequence matching are performed: full title match, main title match, sub-title match, and sequel match, which are inspired by the guidelines in \Cref{ssec:annotation}.
Although this strategy looks simple, it gives a good recall for movie entity recognition; however, it often over-generates candidates resulting a poor precision.
The following shows an example of over-generated candidates, where candidates are shown in bold.

\begin{center}
\textit{\textbf{I do}$_1$ not understand the \textbf{Pompeii}$_2$ \textbf{music videos}$_3$}
\end{center}

\noindent Three candidates are generated from this tweet, ``I do'', ``Pompeii'', and ``music videos''.
Movie titles such as ``I do'' (2012) and ``Pompeii'' (2014) can be problematic for dictionary-based NER systems that do not facilitate any contextual information because these are general expressions and entities.
Our sub-title match allows candidates such as ``music videos'' (from ``NBA Live 2001 - The Music Videos'', 2000), which is not even a movie but included in our gazetteer.
Creating a more fine-grained gazetteer (e.g., IMDB) may resolve issues like this, which we are currently looking into.

\noindent All candidates are automatically tagged with their types (full, main, sub, and sequel), which are later used as features for entity classification.
It might be argued that hashtags alone can be used as a source of candidates; that is, we could generate candidates from only sequences including hashtags.
However, in our data, only 23.6\% of movie entities contain hashtags, and  only 26.3\% of all hashtags occur as part of valid movie entities.
Hence, this approach would fail to identify a significant portion of the movie entities, and also generate too many false-negative candidate entities.


\subsection{Entity classification}
\label{ssec:entity-classification}

For each candidate generated from \Cref{ssec:candidate-identification}, a statistical model is used to classify whether it is a valid entity or not.
Three types of features are used to build the statistical model: orthographic features, \textit{n}-gram features, and syntactic features.
For better generalization, \textit{n}-gram features are supplemented using word embeddings. 
It is worth mentioning that our model does not use any lexical features derived from the candidate itself (e.g., `Hobbit' is not used as a feature), which prevents it from being overfit to any particular set of movies.
Our experiments show that this model gives similar performance on movie entities whose corresponding movies are not in training data compared to ones that are.


\subsubsection{Orthographic features}
\label{sssec:orthographic-features}

Orthographic features give information that can be generally applied to any candidate.
Our model uses the following orthographic features.

\begin{enumerate}
\item \textbf{Capitalization} - Whether the candidate is all capitalized, all decapitalized, or only the first character of each token is capitalized.
\item \textbf{Hashtag} - Whether or not the candidate includes a hashtag.
\item \textbf{Number of tokens} - The total number of tokens in the candidate.
\item \textbf{Title match} - Whether the candidate matches a full title, a main title, a sub-title, or a sequel.
\item \textbf{Numerical time difference} - The difference between the year of the tweet posted and the year of release of the corresponding movie.
\item \textbf{Categorical time difference} - Whether the release date of the corresponding movie is in the past, contemporary, or future with respect to the time of data collection.
\end{enumerate}

\noindent From our experiments, we found that orthographic features could be even more useful than the other features when tweets are written in a proper manner (e.g., proper usage of capitalization or hashtags).
\Cref{ssec:results} gives more detailed analysis of the impact of these features on our system.


\subsubsection{\textit{N}-gram features}
\label{sssec:ngram-features}

\textit{N}-gram features are used to capture the local context of the entities.
Before \textit{n}-gram features are extracted, each candidate is substituted with a generic token, \textsc{\#movie\#}.
\Cref{tbl:ngram-features} shows the \textit{n}-gram feature templates used in our model, where $w_i$ stands for the generic token substituted for the candidate, $w_{i-1}$ and $w_{i+1}$ stand for tokens prior and next to the generic token, and so on.
Note that we also experimented with a bigger window size of tokens, which did not lead to better results in our data.

\begin{table}[htbp!]
\centering
\begin{tabular}{c|l}
\textbf{\textit{N}} & \multicolumn{1}{|c}{\textbf{Template}}\\
\hline\hline
                 1 & $w_{i\pm1}$, $w_{i\pm2}$\\
\hline                 
\multirow{2}{*}{2} & $w_{i-2,i-1}$, $w_{i-2,i+1}$, $w_{i-2,i+2}$,\\
                   & $w_{i-1,i+1}$, $w_{i-1,i+2}$, $w_{i+1,i+2}$\\
\end{tabular} 
\caption{\textit{N}-gram feature templates.  $w_{i,j}$ represents a joined feature between $w_i$ and $w_j$.}
\label{tbl:ngram-features}
\end{table}
\vspace{-2ex}


\subsubsection{Syntactic features}
\label{sssec:syntactic-features}

Syntactic features are derived from dependency trees automatically generated by the dependency parser in ClearNLP~\cite{choi:12a,choi:13a}.
\Cref{tbl:syntactic-features} shows the syntactic feature templates used for our model.

\begin{table}[htbp!]
\centering
\begin{tabular}{l||c|c|c|c}
\multicolumn{1}{c||}{\textbf{Token}} & \textbf{f} & \textbf{m} & \textbf{p} & \textbf{d}\\
\hline\hline
Head       ($h$) & \checkmark & \checkmark & \checkmark & \checkmark\\
Grand-head ($g$) & \checkmark & \checkmark & \checkmark & \checkmark\\
Sibling    ($s$) & \checkmark & \checkmark & \checkmark & \checkmark\\
Dependent  ($d$) & \checkmark & \checkmark & \checkmark & \checkmark\\
\end{tabular} 
\caption{Syntactic feature templates. The f, m, p, and d columns represent a word-form, a lemma, a part-of-speech tag, and a dependency label of the corresponding token, respectively.}
\label{tbl:syntactic-features}
\end{table}

\noindent The generic token, \textsc{\#movie\#}, is usually recognized as a proper noun by the parser, which is proper behavior if it is a valid entity.
We expect that the dependency tree would look more syntactically sound when the generic token is substituted for a valid entity such that features extracted from these trees are distinguishable from features derived from trees that are less syntactically sound. 

\begin{figure*}
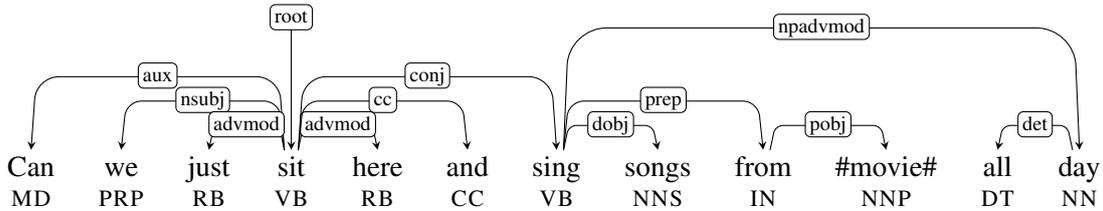

\captionsetup{width=0.9\textwidth}
\begin{dependency}[scale=0.65]
\begin{deptext}[column sep=.4cm]
Can \& we \& just \& sit \& here \& and \& sing  \& songs \& from \& \#movie\# \& all \& day \\
\textsc{md} \& \textsc{prp} \& \textsc{rb} \& \textsc{vb} \& \textsc{rb} \& \textsc{cc} \& \textsc{vb} \& \textsc{nns} \& \textsc{in} \& \textsc{nnp} \& \textsc{dt} \& \textsc{nn} \\
\end{deptext}

\deproot{4}{root}
\depedge{4}{1}{aux}
\depedge{4}{2}{nsubj}
\depedge{4}{3}{advmod}
\depedge{4}{5}{advmod}
\depedge{4}{6}{cc}
\depedge{4}{7}{conj}
\depedge{7}{8}{dobj}
\depedge{7}{9}{prep}
\depedge{7}{12}{npadvmod}
\depedge{9}{10}{pobj}
\depedge{12}{11}{det}

\end{dependency}
\caption{A dependency tree for a tweet, ``Can we just sit here and sing songs from \textsc{\#movie\#} all day'', where the candidate, \textit{Frozen}, is substituted with the generic token, \textsc{\#movie\#}.}
\label{fig:dependency-tree}
\end{figure*}

\begin{table*}[htbp!]
\vspace{1ex}
\centering
\begin{tabular}{c|c}
\textbf{Type} & \textbf{Features} \\ 
\hline\hline
\multirow{2}{*}{Orthographic} & Cap:first\_only, Hashtag:no, Num\_of\_tokens:1, Title\_match:full,\\
                              & Numerical\_time\_diff:0, Categorical\_time\_diff:contemporary \\ 
\hline 
\multirow{3}{*}{\textit{N}-gram} & $w_{i-2}$:songs, $w_{i-1}$:from, $w_{i+1}$:all, $w_{i+2}$:day,\\
                                 & $w_{i-2, i-1}$:songs\_from, $w_{i-2, i+1}$:songs\_all, $w_{i-2, i+2}$:songs\_day,\\
                                 & $w_{i-1, i+1}$:from\_all, $w_{i-1, i+2}$:from\_day, $w_{i+1, i+2}$:all\_day\\ 
\hline 
\multirow{3}{*}{Syntactic} & $h_f$:from, $h_m$:from, $h_p$:\textsc{in}, $h_d$:pobj,\\
                           & $g_f$:sing, $g_m$:sing, $g_p$:\textsc{vb}, $g_d$:dobj\\ 
                           & (\textit{No sibling or dependent features exist for this candidate}.)\\
\end{tabular} 
\caption{Features extracted from the tweet in \Cref{fig:dependency-tree} for the candidate, \textit{\#Frozen}.}
\label{tbl:extracted-features}
\end{table*}


\subsubsection{Supplementary features}
\label{sssec:supplementary-features}

Since our training data is relatively small, \textit{n}-gram features are supplemented with words (or phrases) similar to them in terms of cosine distance as determined by word embeddings.
For each \textit{n}-gram feature, the top 10 most similar words are found from the embeddings and included as additional features.
These supplementary features are weighted by their cosine similarities to the original \textit{n}-gram features.

\begin{table}[htbp!]
\centering
\begin{tabular}{c||c|c|c}
  & \textbf{songs}       & \textbf{awesome}    & \textbf{great film} \\
\hline\hline 
1 & soundtrack  & incredible & train-wreck \\
2 & music       & astounding & masterpiece \\
3 & clips       & excellent  & flick \\
4 & stunts      & amazing    & picture \\
5 & performance & awful      & fine \\
\end{tabular} 
\caption{Examples of the top 5 most similar words extracted from word embeddings.}
\label{tbl:embedding-example} 
\end{table}

\noindent Since our task focuses on movie entities, we build embeddings using our training data (\Cref{ssec:data-split}) combined with two other datasets designed for sentiment analysis on movie reviews~\cite{pang:04a,maas:11a}.\footnote{We used a publicly available tool, \texttt{word2vec}, for building embeddings~\cite{mikolov:13a}.}
The supplementary features are added to improve the generalizability of our model.
Note that we also tried to supplement the \textit{n}-gram features using synonym relations in WordNet~\cite{fellbaum:98a}, which did not show much impact on our system.


\section{Experiments}
\label{sec:experiments}

\begin{table*}[htbp!]
\centering
\begin{tabular}{c|c||c|c||c|c}
\multicolumn{2}{c||}{\textbf{Train}} &\multicolumn{2}{c||}{\textbf{Evaluation 1}} &\multicolumn{2}{c}{\textbf{Evaluation 2}} \\ 
\hline 
\textbf{Movie} & \textbf{Count} & \textbf{Movie} & \textbf{Count} & \textbf{Movie} & \textbf{Count}\\
\hline\hline 
Hobbit           & 194 & Gravity           & 30 & Dallas Buyers Club       & 41\\
Frozen           & 107 & Hobbit            & 27 & Non Stop                 & 24 \\ 
Gravity          & 106 & Frozen            & 26 & Lego Movie               & 21\\
12 Years a Slave &  96 & 12 Years a Slave  & 17 & Lone Survivor            & 20\\
Son of God       &  14 & Lord of the Rings &  4 & Jack Ryan Shadow Recruit &  2\\
\hline\hline
Entities & 667 & Entities & 129 & Entities & 115\\
Movies   &  49 & Movies   &  20 & Movies   &   8\\
\end{tabular} 
\caption{Entity distributions in our training and evaluation sets.  Rows 2 to 6 show the top 5 most frequent movies in each dataset.  The 7th row shows the total number of movie entities in each set.  The last row shows the total number of movies corresponding to these entities.}
\label{tbl:data-split}
\end{table*}

\subsection{Data split}
\label{ssec:data-split}

For our experiments, we divide our corpus into a single training set and two evaluation sets.
The first evaluation set includes only entities corresponding to movies in the training set and the second evaluation set excludes all of those entities.
This allows us to show whether or not our approach is biased to the set of movie entities in our training set.
Since our corpus is not sufficiently large, no development set is separated from the data.
We plan to increase the diversity and the volume of our dataset using crowdsourcing in the near future.

\subsection{Statistical learning}
\label{ssec:algorithm}

Liblinear L2-regularization, L1-loss support vector classification~\cite{hsieh:08a} is used for building the statistical model for entity classification.
We adapt the implementation in ClearNLP,\footnote{\url{http://clearnlp.com}} and use their default learning parameters: $c$ = 0.1, $e$ = 0.1, and $b$ = 0.
Neither hyper-parameter optimization nor feature selection was done for our model; given these small datasets, we did not find further optimization useful since that would bring a higher chance of overfitting the model to this particular data.
Once we annotate more data, we will perform more thorough optimization and try to improve our results with different learning algorithms.

\subsection{Results}
\label{ssec:results}

\begin{table*}[htbp!]
\captionsetup{width=0.85\textwidth}
\centering
\begin{tabular}{l||c|c|c||c|c|c}
\multicolumn{1}{c||}{\multirow{2}{*}{\textbf{Model}}} & \multicolumn{3}{c||}{\textbf{Evaluation  1}} &\multicolumn{3}{c}{\textbf{Evaluation 2}} \\ 
\cline{2-7} & \textbf{P} & \textbf{R} & \textbf{F1} & \textbf{P} & \textbf{R} & \textbf{F1}\\
\hline\hline
Baseline: \Cref{ssec:candidate-identification}  & 14.45    & 96.10          & 25.13         & 13.82          & 96.52         & 24.18\\
Model 1: Baseline + \Cref{sssec:orthographic-features} & 75.00          & 28.13         & 40.09         & 87.60          & 55.65         & 68.09\\
Model 2: Model 1 + \Cref{sssec:ngram-features}         & \textbf{89.01} & 63.28         & 73.97         & \textbf{88.64} & 67.83         & 76.85\\
Model 3: Model 2 + \Cref{sssec:supplementary-features} & 85.44          & 68.75         & 76.19         & 84.16          &\textbf{73.91} & \textbf{78.70}\\
Model 4: Model 3 + \Cref{sssec:syntactic-features}     & 84.76          &\textbf{69.53} &\textbf{76.39} & 79.25          & 73.04         & 76.01\\
\end{tabular} 
\caption{Results of movie entity recognition.  P, R, and F1 represent precision, recall, and F1-score, respectively.  The best result for each category is shown in bold.}
\label{tbl:results}
\end{table*}

\Cref{tbl:results} shows incremental updates of our models.
The baseline results are achieved by running only candidate identification (\Cref{ssec:candidate-identification}).
The results for models 1 to 4 are achieved by running both candidate identification and entity classification using orthographic features (\Cref{sssec:orthographic-features}), \textit{n}-gram features (\Cref{sssec:ngram-features}), supplementary features (\Cref{sssec:supplementary-features}), and syntactic features (\Cref{sssec:syntactic-features}).

As expected, our baseline approach gives very good recall (above 96\%) but poor precision (around 14\%), due to the fact that our gazetteer provides good coverage while causing a large number of false positives for candidate identification.
This implies that the dictionary-based approach itself is not good enough to perform movie entity recognition.
On the other hand, interesting results are derived from Model 1, where orthographic features play a critical role for the 2nd evaluation set but are not as effective for the 1st evaluation set.
From our experiments, we find that these features do not make any noticeable difference when used individually, but give significant improvement when used together (McNemar, $p < 0.0001$).\footnote{No orthographic feature except for the ``number of tokens'' gives any improvement when it is used individually.}

The \textit{n}-gram features used in Model 2 give another significant improvement in both evaluation sets (McNemar, $p < 0.0001$ and $p < 0.01$ for the 1st and 2nd sets, respectively).
The supplementary features used in Model 3 also show improvement in both evaluation sets, but not found to be significant by the McNemar test.
Finally, the syntactic features used in Model 4 give a slight improvement in recall for the 1st evaluation set but hurt both precision and recall for the 2nd evaluation set.
Careful studies with more data need to be conducted to understand why these syntactic features did not show any impact or hurt performance on these datasets.\footnote{We downloaded \newcite{ritter:11a}'s system from their open source project and ran on our data without retraining it, which gave F1-scores of 2.94\% and 20.69\% for the 1st and 2nd evaluation sets, respectively. These results were excluded from \Cref{tbl:results} because it would have not been fair comparison.}

The most interesting aspect about our results is that our final model performs better on the 2nd evaluation set, in which no entity corresponds to any movie in the training set.
This gives strong evidence that our system is completely unbiased to the training data.
\Cref{ssec:analysis-evaluation-sets} describes the different natures of the 1st and 2nd evaluation sets in details.


\section{Discussion}
\label{sec:discussion}

\subsection{Comparison between two evaluation sets}
\label{ssec:analysis-evaluation-sets}

From our error analysis, we notice that tweets in the 2nd evaluation set are much closer to formal English writing than ones in the 1st evaluation set.
This explains why the orthographic features help so much on the 2nd set, but not as much on the 1st set.
Furthermore, movie entities present in the 1st set consist of more general terms (e.g,. \textit{Gravity}, \textit{Frozen}) such that although our training data provides more contextual information for entities in the 1st set, they are harder to recognize.
Nevertheless, the difference between F1-scores achieved for these evaluation sets is not statistically significant, showing that our system performs robustly for movie entities consisting of general terms.

\noindent We also notice that the 2nd evaluation set includes more entities corresponding to movies that are contemporary to the time of our data collection.
Moreover, the 1st set includes more entities composed of a single token; 67.44\% and 22.61\% of entities in the 1st and 2nd evaluation sets are composed of a single token, respectively.
Single-token entities are usually harder to recognize because they tend to use more general terms than multi-token entities.
All these observations show a good direction of how we should split training and evaluation sets for future research.


\subsection{Error analysis}
\label{ssec:error-analysis}

We perform error analysis to identify different types of errors caused by each stage of our system, mainly, candidate identification and entity classification.
\Cref{tbl:error-analysis} shows an example and the proportion of the total errors that each stage causes.

\subsubsection{Errors from candidate identification}

There are three types of errors caused by candidate identification, which together represent 16.4\% of the total errors.
First, entities including typos or non-exact tokens could not be identified by our approach.
For example, an entity such as \textit{Lego Film} is not present in our gazetteer although there exists an entity very similar to it, \textit{Lego Movie}, such that no sequence match is found for this entity.
Second, abbreviated entities are not handled by our approach (e.g, LOTR $\rightarrow$ \textit{Lord of the Rings}).
Third, the maximal sequence matching in \Cref{sssec:sequence-matching} sometimes discards valid entities.
For the 3rd example in \Cref{tbl:error-analysis}, \textit{Heat} should have been identified as a candidate, whereas our approach identifies \textit{in Heat} as a candidate instead because the latter yields a longer sequence.
We could resolve these issues by allowing partial matches~\cite{guo:13a} or combining our dictionary-based approach with statistical approaches~\cite{li:12a}.

\subsubsection{Errors from entity classification}

Most errors in our system are caused in the entity classification stage (83.6\% of the total errors).
Our final model shows much higher precision than recall, which implies that this model generates more false negatives than false positives.
A system with good precision and reasonable recall can be useful for cases where false positives are highly undesirable (e.g., pseudo annotation).
More annotated data with careful feature selection and hyper-parameter optimization may help to improve this stage further.

\begin{table*}[htbp!]
\centering
\begin{tabular}{c|r|l|l}
\textbf{\#} & \multicolumn{1}{c|}{\textbf{\%}} & \multicolumn{1}{c|}{\textbf{Example}} & \multicolumn{1}{c}{\textbf{Note}} \\
\hline\hline
\multirow{3}{*}{1} & \multirow{3}{*}{16.4} & \textbf{NonStop} Unseats \underline{Lego Film}       & \underline{Lego Film} should be recognized.\\ 
                   &                       & I feel like watching \underline{LOTR}                & \underline{LOTR} should be recognized.\\
                   &                       & \textsc{\#user\#} looks like albino \textbf{in \underline{Heat}} & \underline{Heat} should be identified instead of \textbf{in Heat}.\\
\hline
\multirow{2}{*}{2} & \multirow{2}{*}{83.6} & South Park compares her to a \textbf{\#Hobbit} & \textbf{\#Hobbit} is not a movie entity.\\ 
                   &                       & Just got text from \textbf{Nebraska}           & \textbf{Nebraska} is not a movie entity.\\
\end{tabular} 
\caption{Examples of different types of errors caused by each stage.  The \% column shows the proportion of the total errors that each stage causes.}
\label{tbl:error-analysis}
\end{table*}


\section{Related work}
\label{sec:related-work}

It is only recently that natural language processing has been actively applied to social media data.
The inherent difficulties of this domain have been well explored~\cite{foster:11a,dent:11a}.
There has been increasing interest in improving NLP components for this domain.
\newcite{gimpel:11a} and \newcite{owoputi:13a} suggested a new set of part-of-speech tags tailor made for Twitter.
Their \textsc{pos} tagger achieved improved accuracies on tweets, compared to other existing taggers.
 
Named entity recognition is of particular interest in Social Media.
Current state-of-the-art systems perform well on formal texts ~\cite{lin:09a,ratinov:09a,turian:10a}; however, the task becomes much harder in the social media domain.
\newcite{ritter:11a} adapted many NLP components to Twitter data, including a \textsc{pos} tagger, a shallow parser, and an entity chunker.
For movie entities, their dictionary-based approach outperformed their statistical approach, which implies that a new NER system may need to be designed for this type of entities.

\newcite{guo:13a} performed entity linking, a task closely related to NER, on tweets. They developed a system to jointly detect and disambiguate entities in tweets, using structural learning techniques.
Our system is most similar to theirs, in a way that we identify and classify entities. 
However, while they mainly focus on mention detection, we are concerned with building a system to disambiguate a single entity type in a targeted data stream. 
Further, while they use several gazetteer based features in their classification stage, we only use a gazetteer to generate candidates, thus allowing us to refresh our gazetteer without having to retrain our model.

\newcite{li:12a} developed a system for NER in a targeted Twitter stream.
They used dynamic programming to segment each tweet into a set of candidate entities by maximizing what they called a \textit{stickiness function}.
In future work, we aim to incorporate a similar statistical method in our candidate identification stage to reduce our reliance on the gazetteer and improve the precision of this stage. 
Further, they ranked the validity of candidates by constructing a segment graph, and using a random walk model on this graph. 
 
\newcite{xiao:11a} presented a system which combines a KNN classifier and a CRF classifier to label each word in a tweet.
The model is constantly retrained after a pre-determined number of entities are recognized with high confidence. 
As mentioned previously, our system achieves very good precision, and hence, can be used to pseudo annotate new data. 
It would be interesting to evaluate the improvements that can be obtained by using such pseudo-annotated data to occasionally retrain our candidate classification stage.

\newcite{han:11a} and \newcite{kaufmann:10a} examined lexical and syntactic normalization to reduce noise in Twitter data.
We use some of these ideas when we normalize tweets in the first stage in our entity recognition pipeline.


\section{Conclusion}
\label{sec:conclusion}

We provide a rationale for performing careful analysis and developing a dedicated system to recognize targetable named entities in social media data.
In particular, movie entities in tweets are chosen to show the feasibility of our work.
We first create a targeted dataset of tweets containing movie entities, which we make publicly available.
We then present a new system designed for movie entity recognition in tweets.
Our system is evaluated on two datasets, and shows results that are unbiased to our training set.
More importantly, our system can be dynamically adjusted as new movie entities are introduced without having to be retrained.
Although our approach is developed and evaluated only on movie entities, our framework is general enough to be adapted to other types of targetable named entities.

Due to the promising results of this preliminary study, we believe there is great potential for future work.
We plan to increase the size of our corpus with more types of targetable named entities, using crowdsourcing and pseudo annotation.
Moreover, we will improve our gazetteer with more fine-grained sources, and explore statistical approaches for candidate identification.
We also aim to identify features that are more generalizable, and evaluate the adaptability of these features for entity classification.

\bibliographystyle{acl}
\bibliography{references}

\begin{thebibliography}{}

\bibitem[\protect\citename{Bollacker \bgroup et al.\egroup
  }2008]{bollacker:08a}
Kurt Bollacker, Colin Evans, Praveen Paritosh, Tim Sturge, and Jamie Taylor.
\newblock 2008.
\newblock {Freebase: A Collaboratively Created Graph Database For Structuring
  Human Knowledge}.
\newblock In {\em Proceedings of the 2008 ACM SIGMOD International Conference
  on Management of Data}, SIGMOD'08, pages 1247--1250.

\bibitem[\protect\citename{Choi and McCallum}2013]{choi:13a}
Jinho~D. Choi and Andrew McCallum.
\newblock 2013.
\newblock Transition-based dependency parsing with selectional branching.
\newblock In {\em Proceedings of the 51st Annual Meeting of the Association for
  Computational Linguistics}, ACL'13, pages 1052--1062.

\bibitem[\protect\citename{Choi and Palmer}2012]{choi:12a}
Jinho~D. Choi and Martha Palmer.
\newblock 2012.
\newblock {Fast and Robust Part-of-Speech Tagging Using Dynamic Model
  Selection}.
\newblock In {\em Proceedings of the 50th Annual Meeting of the Association for
  Computational Linguistics}, ACL'12, pages 363--367.

\bibitem[\protect\citename{Dent and Paul}2011]{dent:11a}
Kyle Dent and Sharoda Paul.
\newblock 2011.
\newblock {Through the Twitter Glass: Detecting Questions in Micro-text}.
\newblock In {\em AAAI Workshop on Analyzing Microtext}.

\bibitem[\protect\citename{Fellbaum}1998]{fellbaum:98a}
Christiane Fellbaum, editor.
\newblock 1998.
\newblock {\em WordNet: An Electronic Lexical Database}, volume Language,
  Speech and Communications.
\newblock MIT Press.

\bibitem[\protect\citename{Finin \bgroup et al.\egroup }2010]{finin:10a}
Tim Finin, Will Murnane, Anand Karandikar, Nicholas Keller, Justin Martineau,
  and Mark Dredze.
\newblock 2010.
\newblock {Annotating Named Entities in Twitter Data with Crowdsourcing}.
\newblock In {\em Proceedings of the NAACL:HLT Workshop on Creating Speech and
  Language Data with Amazon's Mechanical Turk}, CSLDAMT'10, pages 80--88.

\bibitem[\protect\citename{Foster \bgroup et al.\egroup }2011]{foster:11a}
Jennifer Foster, {\"O}zlem \c{C}etinoglu, Joachim Wagner, Joseph~Le Roux,
  Stephen Hogan, Joakim Nivre, Deirdre Hogan, and Josef van Genabith.
\newblock 2011.
\newblock {\#hardtoparse: POS Tagging and Parsing the Twitterverse}.
\newblock In {\em AAAI Workshop on Analyzing Microtext}, pages 20--25.

\bibitem[\protect\citename{Gattani \bgroup et al.\egroup }2013]{gattani:13a}
Abhishek Gattani, Digvijay~S. Lamba, Nikesh Garera, Mitul Tiwari, Xiaoyong
  Chai, Sanjib Das, Sri Subramaniam, Anand Rajaraman, Venky Harinarayan, and
  AnHai Doan.
\newblock 2013.
\newblock {Entity Extraction, Linking, Classification, and Tagging for Social
  Media: A Wikipedia-Based Approach}.
\newblock {\em Proceedings of the VLDB Endowment}, 6(11):1126--1137.

\bibitem[\protect\citename{Gimpel \bgroup et al.\egroup }2011]{gimpel:11a}
Kevin Gimpel, Nathan Schneider, Brendan O'Connor, Dipanjan Das, Daniel Mills,
  Jacob Eisenstein, Michael Heilman, Dani Yogatama, Jeffrey Flanigan, and
  Noah~A. Smith.
\newblock 2011.
\newblock {Part-of-speech Tagging for Twitter: Annotation, Features, and
  Experiments}.
\newblock In {\em Proceedings of the 49th Annual Meeting of the Association for
  Computational Linguistics: Human Language Technologies}, ACL'11, pages
  42--47.

\bibitem[\protect\citename{Guo \bgroup et al.\egroup }2013]{guo:13a}
Stephen Guo, Ming-Wei Chang, and Emre Kiciman.
\newblock 2013.
\newblock {To Link or Not to Link? A Study on End-to-End Tweet Entity Linking}.
\newblock In {\em Proceedings of the Conference of the North American Chapter
  of the Association for Computational Linguistics on Human Language
  Technology}, NAACL:HLT'13, pages 1020--1030.

\bibitem[\protect\citename{Han and Baldwin}2011]{han:11a}
Bo~Han and Timothy Baldwin.
\newblock 2011.
\newblock {Lexical Normalisation of Short Text Messages: Makn Sens a
  \#Twitter}.
\newblock In {\em Proceedings of the 49th Annual Meeting of the Association for
  Computational Linguistics: Human Language Technologies - Volume 1}, HLT '11,
  pages 368--378. Association for Computational Linguistics.

\bibitem[\protect\citename{Hsieh \bgroup et al.\egroup }2008]{hsieh:08a}
Cho-Jui Hsieh, Kai-Wei Chang, Chih-Jen Lin, S.~Sathiya Keerthi, and
  S.~Sundararajan.
\newblock 2008.
\newblock {A Dual Coordinate Descent Method for Large-scale Linear SVM}.
\newblock In {\em Proceedings of the 25th international conference on Machine
  learning}, ICML'08, pages 408--415.

\bibitem[\protect\citename{Kaufmann}2010]{kaufmann:10a}
M.~Kaufmann.
\newblock 2010.
\newblock Syntactic normalization of twitter messages.
\newblock In {\em International Conference on Natural Language Processing,
  Kharagpur, India.}

\bibitem[\protect\citename{Li \bgroup et al.\egroup }2012]{li:12a}
Chenliang Li, Jianshu Weng, Qi~He, Yuxia Yao, Anwitaman Datta, Aixin Sun, and
  Bu-Sung Lee.
\newblock 2012.
\newblock {TwiNER: Named Entity Recognition in Targeted Twitter Stream}.
\newblock In {\em Proceedings of the 35th International ACM SIGIR Conference on
  Research and Development in Information Retrieval}, pages 721--730.

\bibitem[\protect\citename{Lin and Wu}2009]{lin:09a}
Dekang Lin and Xiaoyun Wu.
\newblock 2009.
\newblock {Phrase Clustering for Discriminative Learning}.
\newblock In {\em Proceedings of the 47th Annual Meeting of the Association for
  Computational Linguistics}, ACL'09, pages 1030--1038.

\bibitem[\protect\citename{Liu \bgroup et al.\egroup }2011]{xiao:11a}
Xiaohua Liu, Shaodian Zhang, Furu Wei, and Ming Zhou.
\newblock 2011.
\newblock {Recognizing Named Entities in Tweets}.
\newblock In {\em Proceedings of the 49th Annual Meeting of the Association for
  Computational Linguistics: Human Language Technologies - Volume 1}, HLT '11,
  pages 359--367. Association for Computational Linguistics.

\bibitem[\protect\citename{Maas \bgroup et al.\egroup }2011]{maas:11a}
Andrew~L. Maas, Raymond~E. Daly, Peter~T. Pham, Dan Huang, Andrew~Y. Ng, and
  Christopher Potts.
\newblock 2011.
\newblock {Learning Word Vectors for Sentiment Analysis}.
\newblock In {\em Proceedings of the 49th Annual Meeting of the Association for
  Computational Linguistics: Human Language Technologies}, ACL:HLT'11, pages
  142--150.

\bibitem[\protect\citename{Mikolov \bgroup et al.\egroup }2013]{mikolov:13a}
Tomas Mikolov, Kai Chen, Greg Corrado, and Jeff Dean.
\newblock 2013.
\newblock {Efficient Estimation of Word Representations in Vector Space}.
\newblock {\em arXiv:1301.3781}.

\bibitem[\protect\citename{Owoputi \bgroup et al.\egroup }2013]{owoputi:13a}
Olutobi Owoputi, Brendan O'Connor, Chris Dyer, Kevin Gimpel, Nathan Schneider,
  and Noah~A. Smith.
\newblock 2013.
\newblock {Improved Part-of-Speech Tagging for Online Conversational Text with
  Word Clusters}.
\newblock In {\em Proceedings of the 2013 Conference of the North American
  Chapter of the Association for Computational Linguistics: Human Language
  Technologies}, NAACL:HLT'13, pages 380--390.

\bibitem[\protect\citename{Pang and Lee}2004]{pang:04a}
Bo~Pang and Lillian Lee.
\newblock 2004.
\newblock A sentimental education: Sentiment analysis using subjectivity
  summarization based on minimum cuts.
\newblock In {\em Proceedings of the 42Nd Annual Meeting on Association for
  Computational Linguistics}, ACL'04, pages 271--278.

\bibitem[\protect\citename{Ratinov and Roth}2009]{ratinov:09a}
Lev Ratinov and Dan Roth.
\newblock 2009.
\newblock {Design Challenges and Misconceptions in Named Entity Recognition}.
\newblock In {\em Proceedings of the Thirteenth Conference on Computational
  Natural Language Learning}, CoNLL'09, pages 147--155.

\bibitem[\protect\citename{Ritter \bgroup et al.\egroup }2011]{ritter:11a}
Alan Ritter, Sam Clark, Mausam, and Oren Etzioni.
\newblock 2011.
\newblock {Named Entity Recognition in Tweets: An Experimental Study}.
\newblock In {\em Proceedings of the Conference on Empirical Methods in Natural
  Language Processing}, EMNLP'11, pages 1524--1534.

\bibitem[\protect\citename{Smith}2013]{smith:13a}
Tom Smith.
\newblock 2013.
\newblock {Twitter Now The Fastest Growing Social Platform In The World}.
\newblock
  http://blog.globalwebindex.net/twitter-now-the-fastest-growing-social-platform-in-the-world/.

\bibitem[\protect\citename{Turian \bgroup et al.\egroup }2010]{turian:10a}
Joseph Turian, Lev-Arie Ratinov, and Yoshua Bengio.
\newblock 2010.
\newblock {Word Representations: A Simple and General Method for
  Semi-Supervised Learning}.
\newblock In {\em Proceedings of the 48th Annual Meeting of the Association for
  Computational Linguistics}, ACL'10, pages 384--394.

\bibitem[\protect\citename{Weiner}1973]{weiner:73a}
Peter Weiner.
\newblock 1973.
\newblock {Linear Pattern Matching Algorithms}.
\newblock In {\em Proceedings of the 14th Annual Symposium on Switching and
  Automata Theory}, SWAT'73, pages 1--11.

\end{thebibliography}

\end{document}